\documentclass{article}
\usepackage[inline]{enumitem}
\usepackage{graphicx}
\usepackage{amsmath}
\usepackage{tabularx}
\usepackage{makecell}
\usepackage[dvipsnames]{xcolor}
\usepackage[font=footnotesize,labelfont=bf]{caption}
\usepackage[draft]{hyperref}
\usepackage{natbib}
\usepackage{placeins}
\usepackage[T1]{fontenc}
\usepackage{CJKutf8}
\usepackage{wrapfig,lipsum,booktabs}
\usepackage[english]{babel}

\usepackage[accepted]{./sty/icml2017_arxiv}
\usepackage{multirow}
\usepackage{dblfloatfix}
\usepackage{slashbox}
\usepackage{makecell}
\usepackage{xcolor,colortbl}

\icmltitlerunning{Recurrent Neural Network-Based Semantic Variational Autoencoder for Sequence-to-Sequence Learning}

\begin{document}
\begin{CJK}{UTF8}{mj}
		
\twocolumn[
\icmltitle{Recurrent Neural Network-Based Semantic Variational Autoencoder \\ 
			for Sequence-to-Sequence Learning}
		
\begin{icmlauthorlist}
	\icmlauthor{Myeongjun Jang}{Ko}
	\icmlauthor{Seungwan Seo}{Ko}
	\icmlauthor{Pilsung Kang}{Ko}
\end{icmlauthorlist}
		
\icmlaffiliation{Ko}{School of Industrial Management Engineering, Korea University, Seoul, South Korea}
\icmlcorrespondingauthor{Pilsung Kang}{pilsung\_kang@korea.ac.kr}
		
\vskip 0.3in
]

\printAffiliationsAndNotice{}  

\begin{abstract}
Sequence-to-sequence (Seq2seq) models have played an important role in the recent success of various natural language processing methods, such as machine translation, text summarization, and speech recognition. However, current Seq2seq models have trouble preserving global latent information from a long sequence of words. Variational autoencoder (VAE) alleviates this problem by learning a continuous semantic space of the input sentence. However, it does not solve the problem completely. In this paper, we propose a new recurrent neural network (RNN)-based Seq2seq model, RNN semantic variational autoencoder (RNN--SVAE), to better capture the global latent information of a sequence of words. To reflect the meaning of words in a sentence properly, without regard to its position within the sentence, we construct a document information vector using the attention information between the final state of the encoder and every prior hidden state. Then, the mean and standard deviation of the continuous semantic space are learned by using this vector to take advantage of the variational method. By using the document information vector to find the semantic space of the sentence, it becomes possible to better capture the global latent feature of the sentence. Experimental results of three natural language tasks (i.e., language modeling, missing word imputation, paraphrase identification) confirm that the proposed RNN--SVAE yields higher performance than two benchmark models.

\textbf{Keywords:} \textit{Sequence-to-sequence learning, Recurrent neural network, Auto-encoder, Variational method, Document information vector, Natural language processing}
\end{abstract}

\section{Introduction}

Sequence-to-sequence (Seq2seq) models \cite{cho2014learning, sutskever2014sequence}, based on recurrent neural networks (RNN), show excellent capability for processing a variable lengths of sequential data. In recent years, these structures have led to the noteworthy development of language models and have played an important role in the development of various tasks of natural language processing (NLP), such as machine translation \cite{bahdanau2014neural, cho2014learning, sutskever2014sequence, ling2015character, luong2015effective, zhao2016deep, lee2016fully, ha2016toward, artetxe2017unsupervised}, machine comprehension \cite{hermann2015teaching, rajpurkar2016squad, yuan2017machine}, text summarization \cite{bahdanau2016end, chan2016listen, nallapati2016sequence}, and speech recognition \cite{graves2014towards, huang2016speaker, chan2016listen, bahdanau2016end}.
\begin{figure*}
	\centering
	\includegraphics[width=1\textwidth]{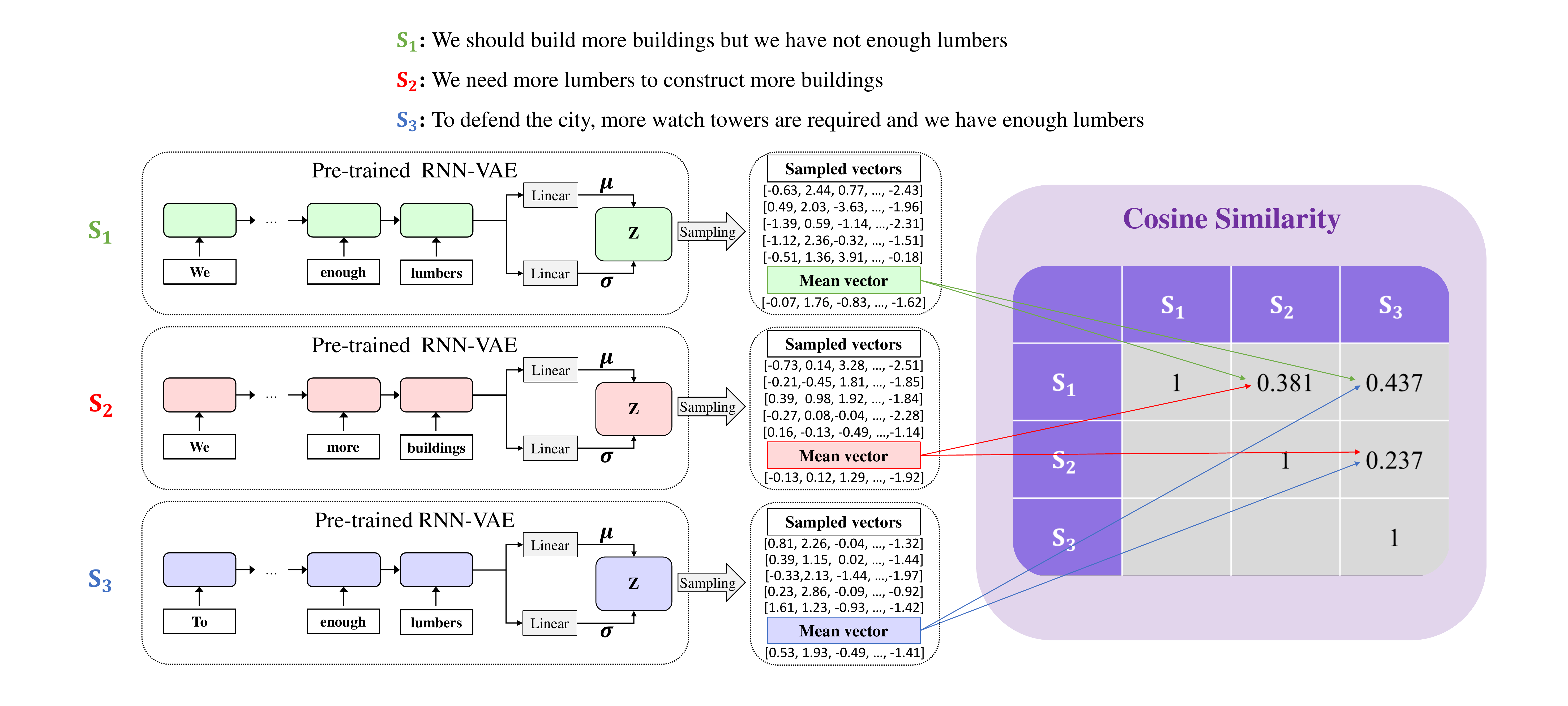}
	\caption{Cosine similarity of sentence information vectors produced by the VAE.}
	\label{fig:VAE_problem}
\end{figure*}
The simplest Seq2Seq structure is the RNN autoencoder (RNN--AE), which receives a sentence as input and returns itself as output \cite{dai2015semi}. Because this model is an unsupervised method that does not require labeled data, it is very easy to obtain training data. Thus, the RNN--AE can be applied to diverse tasks. It has been used to pre-train parameters of a text classification model, achieving better performance than random parameter initialization \cite{dai2015semi}. It also has been used to generate long-length sentences \cite{li2015hierarchical}. Furthermore, it has been applied to not only text data but also to acoustic and video data, which also have sequential information, such as novelty detection of acoustic and video data \cite{marchi2017deep,d2017autoencoder} and representation learning of acoustic data \cite{amiriparian2017sequence}.

Although the RNN--AE shows good performance in many studies, it has limitations. First, because the compressed information of the input sentence is learned as a fixed vector (i.e., encoder final state), there is a high probability that the output will not be good, even for small changes in the vector value. Second, it is not easy to find the global latent feature of the input sentence, owing to the structural features of the RNN that performs the prediction for the next stage of the series \cite{bowman2015generating}. \citet{bowman2015generating} proposed the RNN variational autoencoder (RNN--VAE) model to resolve the above issues by applying a variational inference technique \cite{kingma2013auto, rezende2014stochastic} to the RNN--AE. The RNN--VAE model was successful at moderating the high sensitivity of the RNN--AE to small changes in the final state vector values by learning the information of the input sentence as a probabilistic continuous vector, instead of a fixed vector. The RNN--VAE exhibited a better basic structure than RNN--AE for various NLP tasks, including machine translation \cite{zhang2016variational} and text classification \cite{xu2017variational}

However, the RNN--VAE seems unable to completely solve the problem of RNN-AE, because it seems that it does not capture the global latent feature of the input sentence. With RNN--VAE, the mean and standard deviation of the continuous space of the input sentence information are calculated from the final state of the RNN--AE encoder. Because this final state is updated for each step of the word sequence making up the input sentence, it stores much more information about the last or the beginning parts, if bidirectional RNN is used, of the input sentence, rather than all of the sentence information. Therefore, the continuous space of the input sentence information derived from the final encoder state is hardly a semantic space for preserving the global latent feature. Figure \ref{fig:VAE_problem} shows an example of this RNN--VAE problem with three sentences, S1, S2, and S3. Although S1 and S2 are semantically similar, their syntactic structures are quite different. On the other hand, although S1 and S3 are semantically opposite, they have the same words (enough lumbers) at the end of the sentence. The vector representation of each sentence is the average of sampled vectors from the continuous semantic space of the trained RNN--VAE. We sampled five times for each sentence to reduce bias of the sampled vector and used cosine similarity as the similarity measure between two vector representations. Although S1 is more semantically similar to S2, the cosine similarity between S1 and S3 is higher than that of S1 and S2.

\begin{figure*}
	\centering
	\includegraphics[width=1\textwidth]{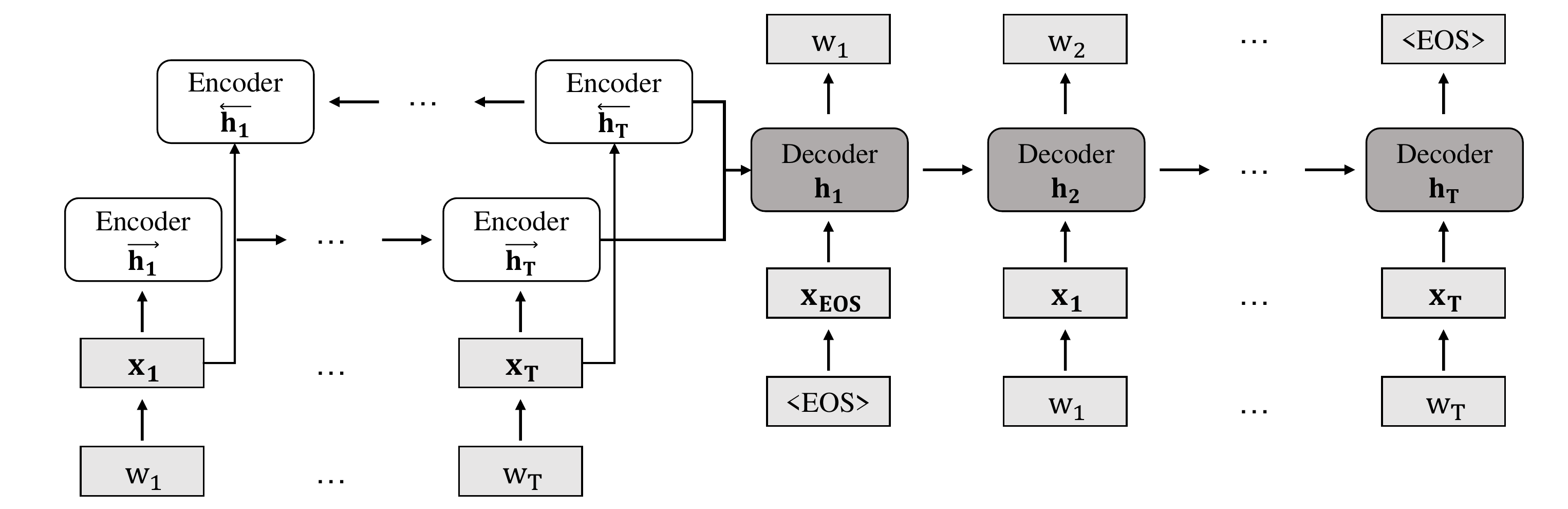}
	\caption{Structure of Seq2Seq AE model with Bi-directional structure.}
	\label{fig:vanilla autoencoder}
\end{figure*}

RNN--AE uses an RNN architecture to construct the encoder and decoder, as shown in Figure \ref{fig:vanilla autoencoder}. The encoder compresses the information of a series of inputs in a sequence (e.g., the words in a sentence), $\textbf{w}=\{w_1,\mathellipsis,w_T\}$, into a fixed vector, $\textbf{v}$,

In this paper, we present RNN--SVAE for overcoming RNN--VAE limitations by generating a document information vector to capture the global latent feature of the input sentence. The document information vector consists of the word weights of a linear combination most correctly representing the paragraph vector using word vectors in the embedding space. This document information vector is combined with the final encoder state. The RNN--SVAE is trained based on the combined vector to find an appropriate continuous space of the input sentence. RNN--SVAE's effectiveness is verified by comparing its performance to that of RNN--AE and RNN-VAE, using three tasks: language modeling, missing word imputation, and paraphrase identification.

The rest of this paper is organized as follows. In Section 2, we briefly review past research on the autoencoder structure and demonstrate the methodologies used in this study. In Section 3, we catalogue the architecture of RNN--SVAE. In Section 4, experimental settings of each task are described, followed by results and discussion. Finally, in Section 5, we conclude our current work with some future research directions.

\section{Background}
\subsection{RNN--AE}

The AE, first introduced by \citet{rumelhart1985learning}, is a neural network-based unsupervised learning algorithm that has been employed for various tasks, including feature representation, anomaly detection, and transfer learning \cite{baldi2012autoencoders, bengio2013representation, zhu2016deep, sakurada2014anomaly, chen2017outlier, lyudchik2016outlier, zhuang2015supervised, deng2013sparse}. Input and output are the same in the AE structure. Thus, the AE’s learning objective is to approximate the output to the input as closely as possible. The preceding part, compressing the information of the input vector to the latent vector, is called the $encoder$, and the following part, reconstructing the information from the latent vector to the output, is called the $decoder$.

\begin{figure*}
	\centering
	\includegraphics[width=1\textwidth]{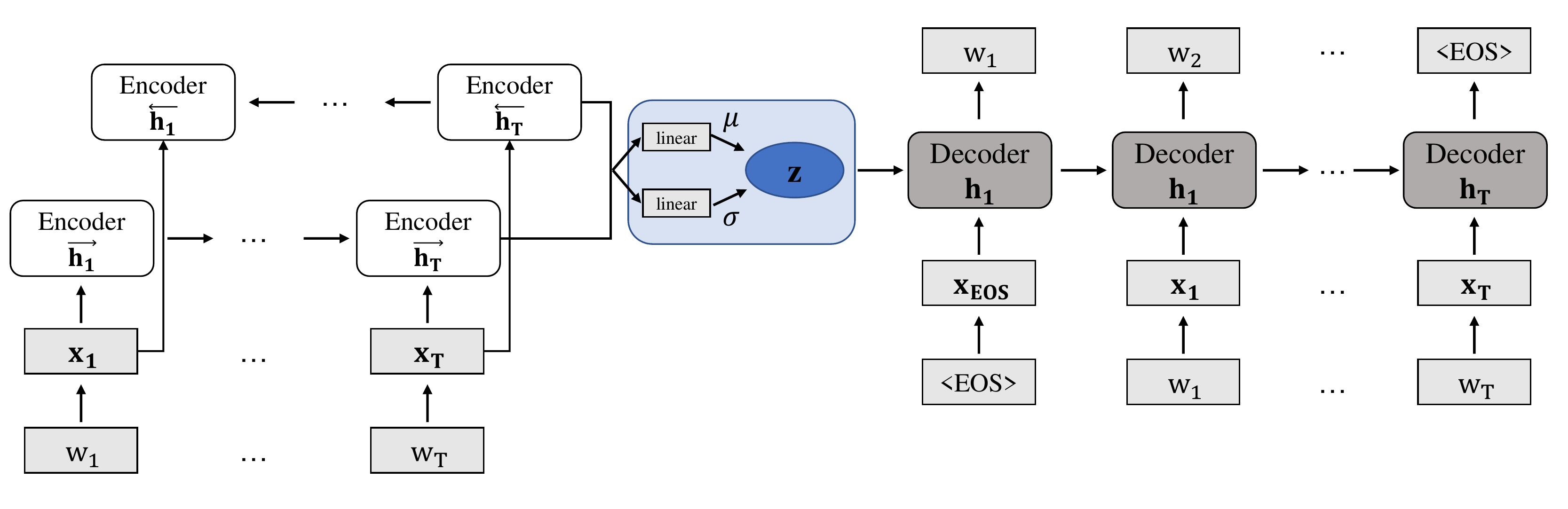}
	\caption{Structure of RNN--VAE model with Bi-directional structure}
	\label{fig:VAE}
\end{figure*}
The AE, first introduced by \citet{rumelhart1985learning}, is a neural network-based unsupervised learning algorithm that has been employed for various tasks, including feature representation, anomaly detection, and transfer learning \cite{baldi2012autoencoders, bengio2013representation, zhu2016deep, sakurada2014anomaly, chen2017outlier, lyudchik2016outlier, zhuang2015supervised, deng2013sparse}. Input and output are the same in the AE structure. Thus, the AE’s learning objective is to approximate the output to the input as closely as possible. The preceding part, compressing the information of the input vector to the latent vector, is called the $encoder$, and the following part, reconstructing the information from the latent vector to the output, is called the $decoder$.

\begin{gather}
\textbf{h}_i = f(\textbf{x}_i,\textbf{h}_{i-1}),\\
\textbf{v} = q( \{ \textbf{h}_1,\mathellipsis, \textbf{h}_T \}),
\end{gather}
where $w_i$ is the $i^{th}$ input word, $\textbf{x}_i$ is the input vector of the $w_i$, and $\textbf{h}_i$ is the hidden state of the $i^{th}$ sequence. $f$ and $q$ are nonlinear functions, where $q( \{ \textbf{h}_1,\mathellipsis, \textbf{h}_T \})=\textbf{h}_T$. The decoder is trained to maximize the conditional probability of predicting the next word, $\hat{y}_{t'}$, given a fixed vector, $\textbf{v}$, and previously predicted words, $\{\hat{y}_1,\mathellipsis,\hat{y}_{t'-1}\}$. Thus, the purpose of the decoder is to maximize the probability of predicting the target sequence, $\textbf{y}=\{y_1,\cdots,y_{t'}\}$,
\begin{gather}
p(\hat{\textbf{y}}) = \prod^{T'}_{t=1}p(\hat{y}_t| \{ \hat{y}_1, \mathellipsis,\hat{y}_{t-1} \} ,\textbf{v}).
\end{gather}
Because the objective is to precisely reconstruct the input, $\textbf{y}$ is identical to $\textbf{x}$ in the RNN--AE. The conditional probability of RNN structure at time, $t$, is defined as
\begin{gather}
p(\hat{y}_t| \{ \hat{y}_1, \mathellipsis, \hat{y}_{t-1} \}, \textbf{v}) = g(\hat{y}_{t-1},\textbf{s}_t,\textbf{v}),
\end{gather}
where $\textbf{s}_t$ and $g$ denote the hidden state of the decoder at time, $t$, and a non-linear function, respectively.

\subsection{RNN--VAE}

RNN--VAE is a generative model that improves the RNN--AE to capture the global feature of the input sentence. RNN--VAE replaces the deterministic function, $q( \{ \textbf{h}_1,\mathellipsis, \textbf{h}_T \})=\textbf{h}_T$, of RNN—AE with the posterior recognition model, $q(\textbf{z}|\textbf{x})$, which compresses the information of input sentence, $\textbf{x}$, into a probabilistic distribution. The parameters, $\boldsymbol{\mu}$ and $\boldsymbol{\sigma}$, determining $q(\textbf{z}|\textbf{x})$, are calculated as a linear transformation of the encoder output. Thus, RNN--VAE is a model that learns the compressed information of the input sentence as a region of latent space, rather than as a single point. The structure of RNN--VAE model is shown in Figure \ref{fig:VAE}

If the RNN--VAE is trained only with the RNN--AE's reconstruction objective, it would encode the input sentence as an isolated point which means that it makes the variance of $q(\textbf{z}|\textbf{x})$ very small \cite{bowman2015generating}. To deal with this problem, in addition to the reconstruction objective, the RNN--VAE has another objective that approximates the posterior distribution, $q(\textbf{z}|\textbf{x})$, to the prior distribution, $p(\textbf{z})$. This is generally a standard Gaussian distribution, ($\mu=\stackrel{\rightarrow}{0},~\sigma=\stackrel{\rightarrow}{1}$). The Kullback-Leibler divergence (KLD) is used to compute the difference between the two distributions. Thus, the objective of RNN--VAE is defined as
\begin{gather}
	\scalebox{0.89}{$
	\begin{aligned}
	\mathcal{L}(\theta :\textbf{x}) = -KLD(q_{\theta}(\textbf{z}|\textbf{x})|p(\textbf{z})) + E_{q_\theta(\textbf{z}|\textbf{x})}[logp_\theta(\textbf{x}|\textbf{z})],\label{eq:VAE_obj}
	\end{aligned}$}
\end{gather}
where $\theta$ is the model parameter (i.e., $\mu$ and $\sigma$ of Gaussian distribution) in the RNN--VAE. This objective allows the RNN--VAE to decode output at every point in the continuous space, having high probability under the prior distribution.

\subsection{Paragraph Vector}
Paragraph vector \cite{le2014distributed} has been widely used to represent a paragraph using an arbitrary number of words into a fixed low-dimensional continuous vector to overcome the limitations of the bag-of-words (BoW) method. There are two main ways to learn the paragraph vector: the paragraph vector with distributed memory (PV-DM) method and the paragraph vector with distributed BoW (PV-DBOW) method. The PV-DM method, which considers the order of word sequence, has a similar model structure to continuous BoW (CBOW) of the Word2Vec model. This model takes the paragraph token vector, $\textbf{p}$, and word vectors, $\textbf{x}_i, \mathellipsis, \textbf{x}_{i+(t-1)}$, to predict the next word, $\textbf{x}_{i+t}$, when the sliding window size is set to $t$. Thus, the paragraph vector is trained to maximize the probability of its appearance with the words contained in the sliding window of the paragraph. In the PV-DBOW method, the words included in the fixed window are arbitrarily sampled from those constituting the paragraph. This model takes the paragraph vector as input and predicts the sampled words. Therefore, it does not consider the order of the paragraph’s word sequence. Both methods define the probability that a paragraph token and a word token appear together, using the dot product between the vectors of each token. Therefore, the paragraph vector is located close to the word vectors within the paragraph from the semantic embedding.

\begin{figure*}
	\centering
	\includegraphics[width=1\textwidth]{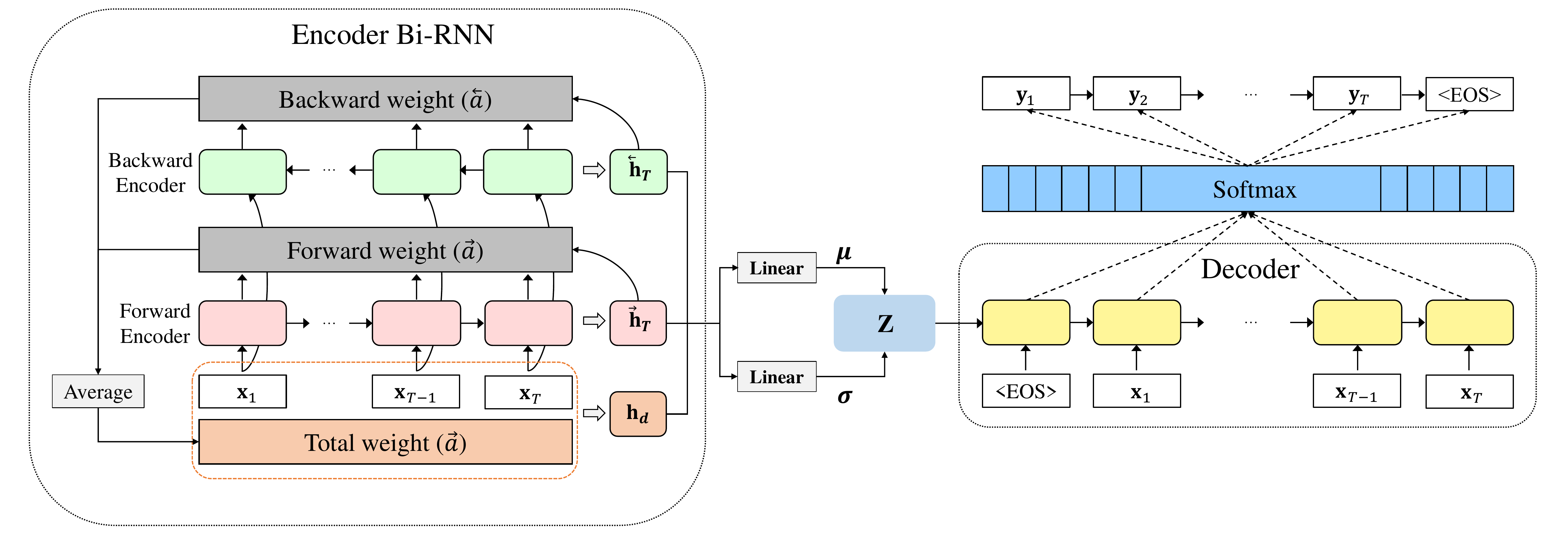}
	\caption{Structure of SVAE model}
	\label{fig:SVAE}
\end{figure*}
\subsection{Attention Mechanism}
The attention mechanism \cite{luong2015effective,bahdanau2014neural}, recently recognized for its effectiveness, is widely used for image captioning \cite{xu2015show}, tree parsing \cite{vinyals2015grammar}, question answering \cite{hermann2015teaching}, and machine translation. The main problem of vanilla RNN is that it hardly preserves the information of the words from the front of sentence when the input sentence becomes longer. This is because it only uses the last hidden state of the encoder. Although the long short-term memory (LSTM, \cite{hochreiter1997long}) or gated recurrent unit (GRU, \cite{cho2014properties}) tends to alleviate the problem, they still have trouble preserving the well-balanced semantic information of a sentence, regardless of the word appearance sequence. An attention mechanism solves this problem by using a weighted combination of total hidden states (i.e., context vector) and the last hidden state for each decoding step. The weights of the context vector can be regarded as the importance of input sentence words in the corresponding step.

\section{Model Structure}
In this study, we propose the RNN semantic variational autoencoder (RNN--SVAE) which represents the global latent feature of an input sentence better than RNN--VAE. As shown in Figure \ref{fig:SVAE}, RNN--SVAE integrates the final hidden state and the document information vector, based on the attention vectors of bi-directional RNN (bi-RNN) hidden states, before estimating the parameters of Gaussian distribution. Because every word in the input sentence is equally considered in the document information vector, the RNN--SVAE can preserve the global latent feature better than the RNN--VAE, which has highly skewed information toward the latter words. Additionally, because the document information vector is computed by aggregating the attention vectors, model-training is not required to separately learn the document information vector; it is learned simultaneously with the hidden state during RNN training.

\subsection{Document Information Vector}
For both PV-DM and PV--CBOW methods, a paragraph vector is placed near the word vectors constituting it because a $d$-dimensional paragraph vector is trained to maximize the dot product of the $d$-dimensional word vectors in the paragraph. This implies that a linear combination of $d$ linearly independent word vectors can accurately reconstruct the paragraph vector. Furthermore, because the paragraph vector has a high similarity to the vectors of words constituting the paragraph, it is possible to approximate the paragraph vector using the embedding vectors of its words, as follows.
\begin{gather}
\textbf{v}_d \cong \sum_{i=1}^T a_i\textbf{x}_i,
\end{gather}
where $\textbf{x}_i$ and $a_i$ denote the $i^{th}$ word vector and its linear combination weight, respectively. $\textbf{v}_d$ is the paragraph vector.

Whereas PV-DM and PV--CBOW explicitly learn the paragraph vector during model training, the proposed document information vector computes it implicitly using information obtained during Seq2seq model training. Because the last hidden state, ($\textbf{h}_T$), of the encoder, is a vector containing the sequential information of the input sentence, we compute the weight, $a_i$, using the relationship between the $\textbf{h}_T$ and the $i^{th}$ hidden state, ($\textbf{h}_i$). Many past studies used their dot product as a similarity measure \cite{karpathy2014deep,karpathy2015deep}. We instead use the normalized value of the dot product between $\textbf{h}_T$ and $\textbf{h}_i$ as the $a_i$,
\begin{gather}
a_i = \frac{e_i}{\sum_{k=1}^T e_k}, ~~~~~ \textrm{where} ~~~ e_i = \textbf{h}_i \cdot \textbf{h}_T.
\end{gather}
It is possible to use many other alignment models that are proposed by \citet{luong2015effective} or \citet{bahdanau2014neural}. However, we used a simple normalized dot product to focus on the effectiveness of document vector itself.

Using the standard RNN structure tends to give a larger weight to the words at the end of the input sentence. The closer $i$ is to $T$, the more similar $\textbf{h}_i$ is to $\textbf{h}_T$. To solve this problem, we use bi-RNN \cite{schuster1997bidirectional} to take the average of the forward weight, ($\stackrel{\rightarrow}{a}$), and the backward weight, ($\stackrel{\leftarrow}{a}$),
\begin{gather}
\scalebox{0.89}{$
	\begin{aligned}
	\stackrel{\leftarrow}{a_i} = \frac{\stackrel{\leftarrow}{\textbf{h}_i} \cdot \stackrel{\leftarrow}{\textbf{h}_T}}{\Sigma^T_{k=1}\stackrel{\leftarrow}{\textbf{\textbf{h}}_k} \cdot \stackrel{\leftarrow}{\textbf{h}_T}}, ~~
	\stackrel{\rightarrow}{a_i} = \frac{\stackrel{\rightarrow}{\textbf{h}_i} \cdot \stackrel{\rightarrow}{\textbf{h}_T}}{\Sigma^T_{\textbf{k}=1}\stackrel{\rightarrow}{\textbf{h}_k} \cdot \stackrel{\rightarrow}{\textbf{h}_T}}, ~~
	\bar{a_i} = \frac{\stackrel{\rightarrow}{a_i} + \stackrel{\leftarrow}{a_i}}{2},
	\end{aligned}$}
\end{gather}
where $\stackrel{\rightarrow}{\textbf{h}_i}$ and $\stackrel{\leftarrow}{\textbf{h}_i}$ is the forward and backward hidden states at the $i^{th}$ word, respectively. Finally, we compute the document information vector by combining the total weight, ($\bar{a}$), and the word sequence, $\textbf{x}=\{\textbf{x}_1,\mathellipsis,\textbf{x}_T\}$, of the input sentence,
\begin{gather}
\textbf{h}_d = \sum^T_{i=1}\bar{a_i}\textbf{x}_i,
\end{gather}
where the $\textbf{h}_d$ is the document information vector of the input sentence. Contrary to the paragraph vector, which should be trained separately from the RNN model to learn the sentence vector, the proposed document information vector can be computed simultaneously using the learned parameters of the RNN model. Hence, unlike the paragraph vector, it is not necessary to learn the sentence vector for each new sentence.

\begin{table*}[t!]
	\begin{center}
		\caption{Number of sentences for each data set} \label{table2.Number of sentences for each data set}%
		\renewcommand{\arraystretch}{1.3}
		\small{
			\centering{\setlength\tabcolsep{8pt}
				\begin{tabular}{c|c|c|c|c|c}
					\hline
					&
					News Crawl'13 & News Crawl'14 & News Crawl'15 & News Crawl'16 & TED Talk \\ 
					\hline
					Train &  2,500,000&  - &  - &  - &   - \\ 
					\hline
					Test & 15,000 & 15,000 & 15,000 & 15,000 & 15,000 \\ 
					\hline
		\end{tabular}}}
	\end{center}
\end{table*}

\subsection{RNN--SVAE}
The structure of RNN--SVAE model is created by adding the document information vector to the RNN--VAE model. The overall structure of the proposed model is summarized in Figure \ref{fig:SVAE}. We construct the final state of the encoder by concatenating the forward final state ($\stackrel{\rightarrow}{\textbf{h}_T}$), backward final state ($\stackrel{\leftarrow}{\textbf{h}_T}$), and document information vector ($\textbf{h}_d$) as follows,
\begin{gather}
\textbf{h}_L = [\stackrel{\rightarrow}{\textbf{h}_T} ~ ; ~ \stackrel{\leftarrow}{\textbf{h}_T} ~ ; ~ \textbf{h}_d]
\end{gather}
Next, the mean, ($\boldsymbol{\mu}$), and the standard deviation, ($\boldsymbol{\sigma}$), vectors of the continuous semantic space is calculated from the encoder’s last state ($\textbf{h}_L$) via linear transformation. These vectors have the same dimension as the global latent vector, ($\textbf{z}$). Finally, we sample the global latent vector, which functions as the semantic vector, of the input sentence and is used as an input vector to the decoder, from the continuous semantic space.
\begin{gather}
\boldsymbol{\mu} = \textbf{W}_\mu \textbf{h}_L + \textbf{b}_{\mu}, ~~~ \boldsymbol{\sigma} = \textbf{W}_\sigma \textbf{h}_L + \textbf{b}_\sigma, \\
\textbf{z} \sim Gaussian(\stackrel{\rightarrow}{0},\stackrel{\rightarrow}{1}),
\end{gather}
where $\textbf{W}_\mu$ and $\textbf{b}_\mu$ are the weight and bias for $\boldsymbol{\mu}$, respectively, whereas, $\textbf{W}_\sigma$ and $\textbf{b}_\sigma$ are the weight and bias, respectively, for $\boldsymbol{\sigma}$.

Similar to RNN--VAE, the RNN--SVAE's cost function reflects two objectives, as shown in Eq. (\ref{eq:VAE_obj}).
The first objective is to closely approximate the posterior distribution, $q(\textbf{z}|\textbf{x})$, with the parameters, $\boldsymbol{\mu}$ and $\boldsymbol{\sigma}$, to the prior distribution, $p(\textbf{z})$, which is the standard Gaussian distribution. To do so, the $KLD$, having the standard Gaussian distribution, should be minimized. The second objective is to maximize the conditional probability, $p(y_1,\mathellipsis,y_{T'}|\textbf{x}_1,\cdots,\textbf{x}_T)$, like in the general Seq2seq model. $\textbf{w}=\{w_1,\mathellipsis,w_T\}$ is the word sequence of the input sentence, and $\textbf{y}=\{y_1,\mathellipsis,y_{T'}\}$ is the output sequence. Because the RNN--SVAE model is an autoencoder structure, $\textbf{w}$ is identical to $\textbf{y}$.

\section{Experiments}
During our experiments, we verified the RNN--SVAE with the three tasks: language modeling, missing word imputation, and paraphrase identification. As baseline models, RNN--AE and RNN--VAE were also used. Evaluation and comparison were both conducted quantitatively with the standard evaluation metrics, and qualitatively by exploiting the output examples of the three models. 

\subsection{Language Modeling}
To evaluate the fundamental ability of RNN-SVAE as an autoencoder, language modeling was tested first. 

\subsubsection{Data Set and Preprocessing}

In this study, we used the News Crawl data of WMT' 17\footnote{http://www.statmt.org/wmt17/translation-task.html}, English monolingual corpus, and TED Talk data of WIT3\footnote{https://wit3.fbk.eu/} \cite{cettolo2012wit3}. The News Crawl'13 dataset was used to train the RNN models, whereas all datasets were used to test the models. For the language modeling task, it is common to exclude very long sentences (i.e., longer than 30 to 50 words) to accelerate training \cite{bahdanau2014neural, artetxe2017unsupervised}. Therefore, we only used sentences shorter than 40 words for computational efficiency. For the training dataset, we randomly sampled 2,500,000 sentences from the News Crawl'13 dataset. As test dataset, News Crawl'13, News Crawl'14, News Crawl'15, and News Crawl'16 datasets of WMT' 17 English monolingual corpus and the TED Talk dataset were used. For each dataset, 15,000 sentences were randomly sampled. For the test data of News Crawl'13 dataset, the sampled sentences in the training dataset were excluded when sampling the test dataset. The number of training and test sentences in each dataset is summarized in Table \ref{table2.Number of sentences for each data set}.

Prior to training the model, we performed tokenization\footnote{We used RegExpTokenizer from the NLTK package. (http://www.nltk.org/api/nltk.html)} after removing punctuation marks and converting uppercase letters to lowercase letters for all sentences. Following tokenization, we pre-trained the word vectors by using the skip-gram model \cite{mikolov2013distributed}. Words that appeared fewer than seven times in the training dataset were replaced with the ``UNK'' token (i.e., unknown word). We set the dimension of word vector to 100, the window size of the skip-gram model to 5, and the negative sampling parameter, $k$, to 5. Word vector training was done for 10 epochs. Thus, including ``UNK'' and ``EOS'' (i.e., end-of-sentence) token, 91,897 unique words were trained.

\begin{table*}[t!]
	\begin{center}
		\caption{BLEU score of each model for the language modeling task} \label{table3.BLEU score of Language modeling result}%
		\renewcommand{\arraystretch}{1.3}
		\small{
			\centering{ \setlength\tabcolsep{5pt}
				\begin{tabular}{c|c|c|c|c|c}
					\hline
					\backslashbox{Model}{Data} & 
					News Crawl'13 & News Crawl'14 & News Crawl'15 & News Crawl'16 &
					TED Talk \\ 
					\hline
					RNN--AE &  17.55 &  20.21 &  19.98  &  20.55  & 24.84 \\ 
					\hline
					RNN--VAE & 37.51 & 38.62 & 39.41 & 38.98 & 45.49\\ 
					\hline
					RNN--SVAE & \textbf{$$41.68$$} & \textbf{$$41.89$$} & \textbf{$$43.33$$} & \textbf{$$43.07$$} & \textbf{$$49.32$$}\\ 
					\hline
		\end{tabular}}}
	\end{center}
\end{table*}

\begin{figure*}[t!]%
	\centering{
		\begin{tabular}{ccc}
			\includegraphics[width=0.32\textwidth]{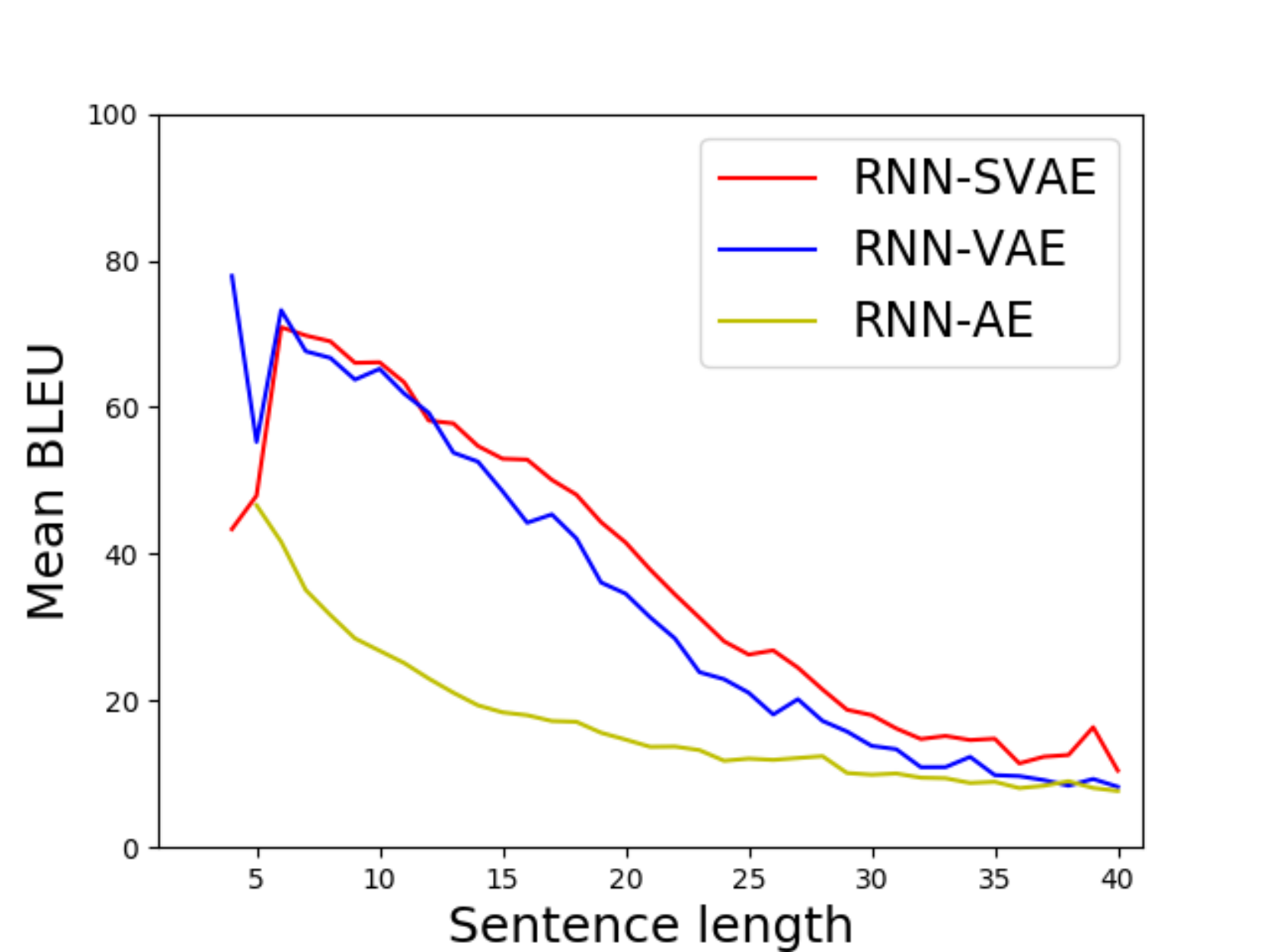} &
			\includegraphics[width=0.32\textwidth]{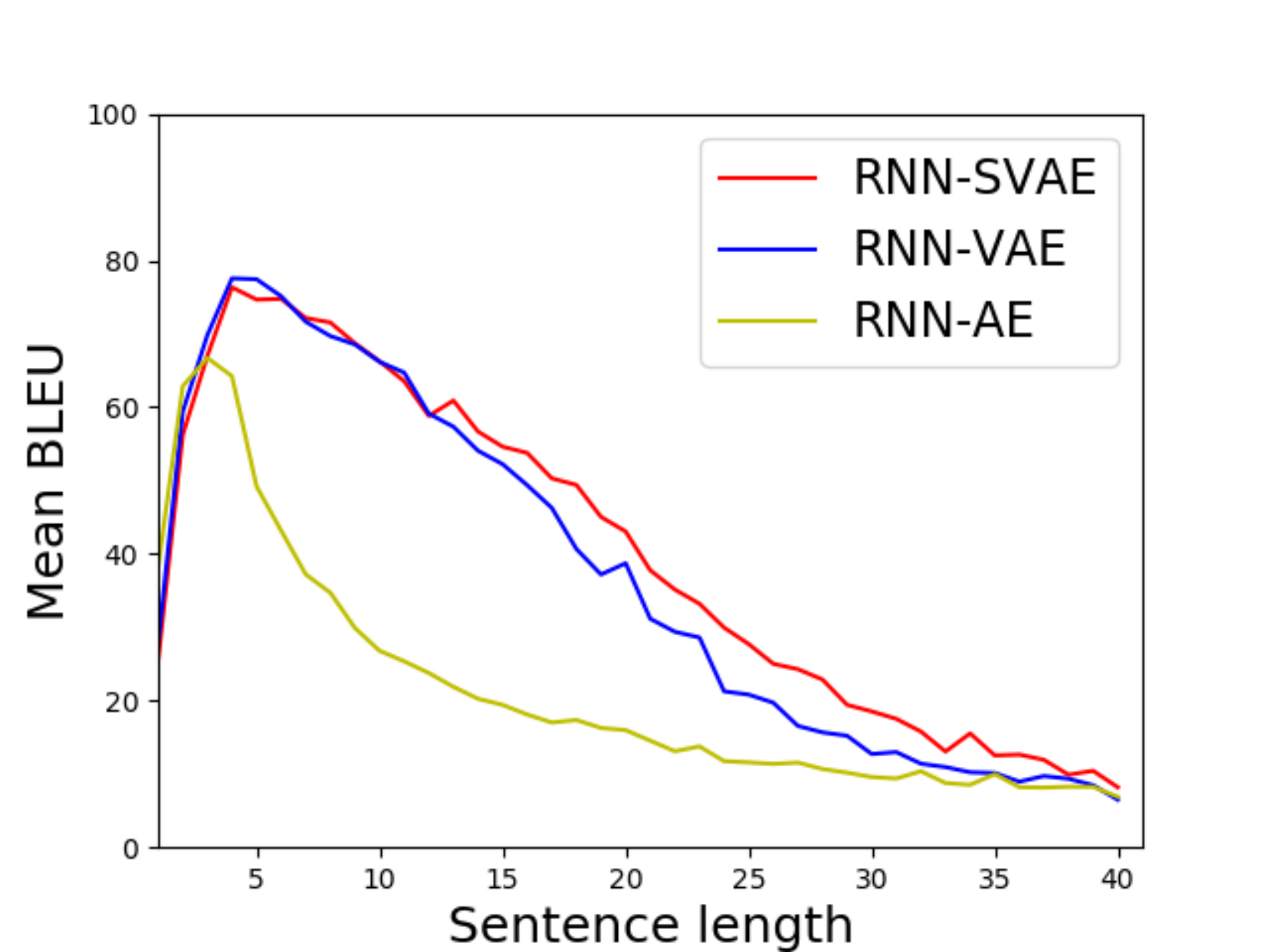} &
			\includegraphics[width=0.32\textwidth]{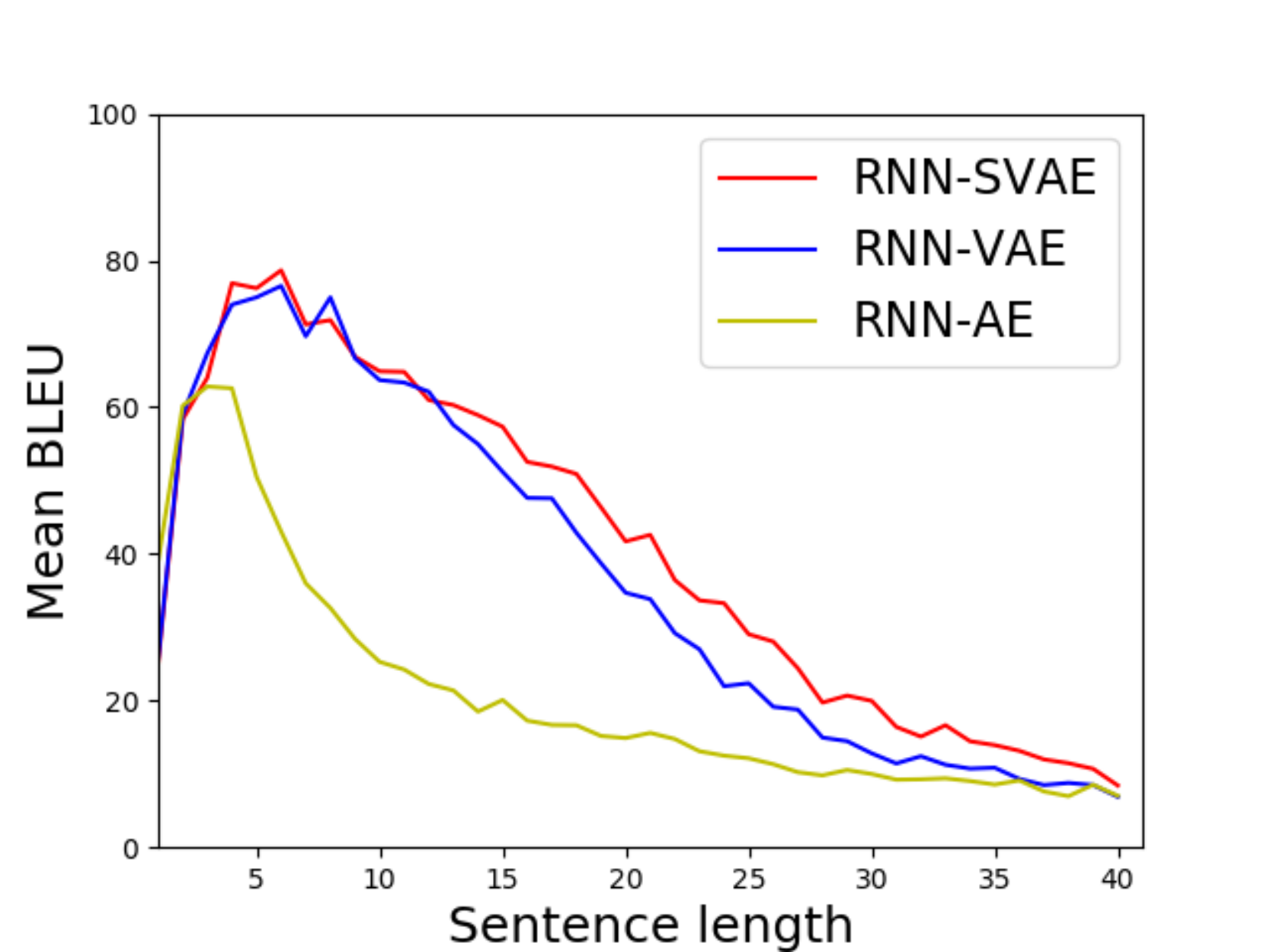} \\
			\footnotesize{(a) News Crawl'13} & \footnotesize{(b) News Crawl'14} & \footnotesize{(c) News Crawl'15}\\ %
			\includegraphics[width=0.32\textwidth]{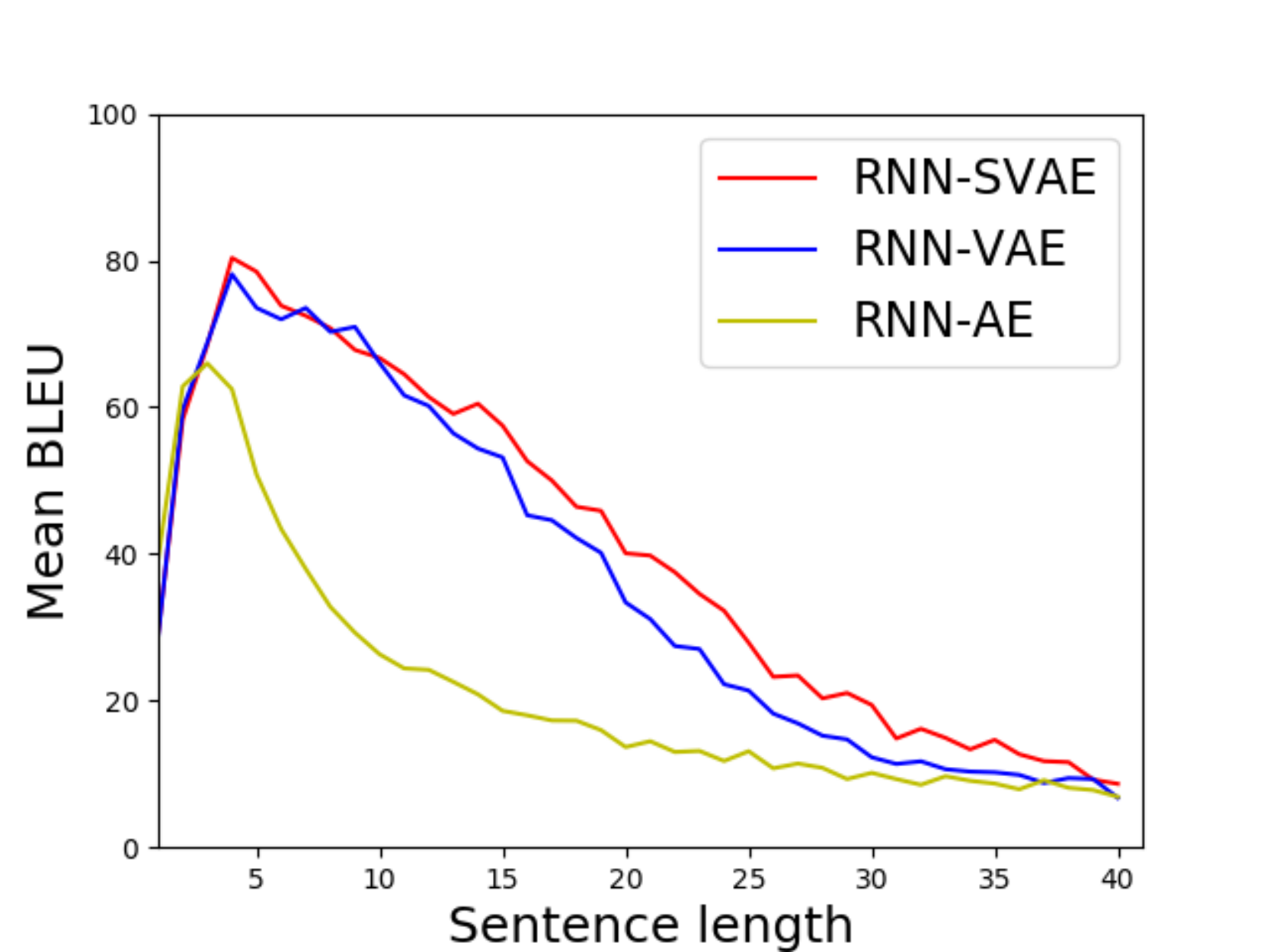} &
			\includegraphics[width=0.32\textwidth]{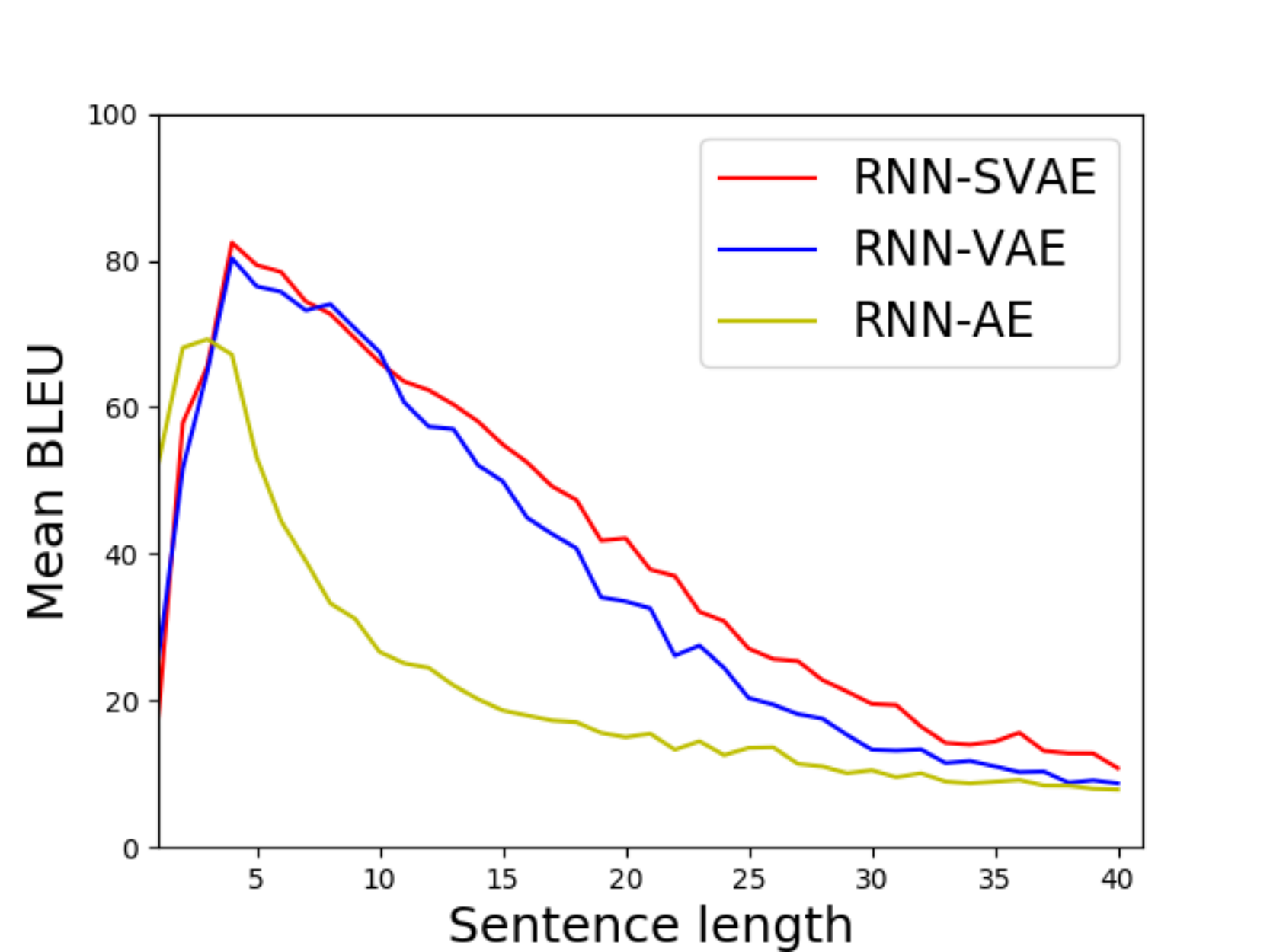} \\
			\footnotesize{(d) News Crawl'16} & \footnotesize{(e) TED Talk} \\ %
	\end{tabular}} %
	\caption{BLEU scores for each model according to the sentence length.}
	\label{Fig:BLEU}%
\end{figure*}

\subsubsection{Model Training and Inference}
Because the RNN--SVAE model is rooted on vanilla RNN, it is possible to use any type of RNN cell (e.g., basic RNN cell, LSTM cell, GRU cell) \cite{cho2014properties}. We used the GRU cell that solves the gradient vanishing problem of the basic RNN cell. It also has fewer parameters than the LSTM cell. The RNN--SVAE encoder has a bi-directional RNN structure. The forward and backward RNNs of the encoder each consist of 300 hidden units. The global latent vector and hidden states of the decoder also consist of 300 units.

For fair comparison, the baseline models are designed with the same structure as the RNN-SVAE model. The RNN--VAE model also used the GRU cell and had a bi-RNN structure with 300 hidden units. Its global latent vector and decoder hidden units were composed of 300 units, as in the RNN--SVAE model. The RNN--AE model also used the GRU cell with bi-RNN structure and had the encoder and decoder with 300 hidden units, as in the RNN--VAE and RNN--SVAE models.

The three models were all trained under the same condition. We initialized their parameters using the Xavier initialization \cite{glorot2010understanding}. We used the Adam optimizer \cite{kingma2014adam} for training. We trained the models for 30 epochs. Gradient computation and weight update were done with the mini-batch size of 512. The learning rate was set to 0.001 for the first 10 epochs and to 0.0001 for the remaining 20 epochs.

After model training, beam search was used to obtain the output maximizing conditional probability at the inference phase. We set the beam size to 7 and the maximum length of output to 40. For generative models, such as RNN--VAE and RNN--SVAE, an average of five samples was used as the input vector of the decoder to reduce the bias of sampled global latent vector.
 
 \subsubsection{Results}
As a performance measure for language modeling, the BLEU score, commonly used for machine translation, was used \cite{papineni2002bleu}. Although there exists other ``teacher forcing'' metrics, such as negative log likelihood and perplexity, these metrics are insufficient in evaluating whether the semantic space or vector, i.e. the output of the encoder, reflects the global latent feature of the input sentence because the target token at every time step is provided for those ``teacher forcing'' metrics. As a result, we used BLEU score as a performance measure that target tokens are not provided in each decoding step. The results of each model are summarized in Table \ref{table3.BLEU score of Language modeling result}. For all five test data sets, the proposed RNN--SVAE significantly outperformed the benchmark models. The BLEU scores of RNN--SVAE were almost as twice those of the RNN--AE. Compared to RNN--VAE, RNN--SVAE improved the BLEU score by at least 3.27 (News Crawl'14) and at most 4.17 (News Crawl'13). The relative BLEU improvements of RNN--SVAE against RNN--VAE were between 8.42\% and 11.12\%.

Table \ref{table4.A few examples of language modeling outputs produced by VAE and SVAE} shows examples of the language modeling task\footnote{We used smoothing function in NLTK package, because the BLEU score was too optimistic}. Examples of RNN--AE are not given, because its language modeling performance was significantly worse than those of RNN--VAE and RNN--SVAE. Highlighted parts, in gray, show the words exactly matched with the ground truth. In the case of RNN--VAE, both the beginning and the end of sentences fit well with the ground truth. However, it seemed to have difficulty generating the parts in the middle of sentences correctly. However, RNN--SVAE succeeded in generating the entire sentences. 

Whereas the word sequences generated by the RNN--SVAE are not the same as the ground truth, the chosen words are semantically very similar to those in the ground truth. As shown in Table \ref{table5.Examples of wrong words but have high similarity with ground truth}. When the word ``nodes'' is replaced by ``fibres'' in the first example sentence, it is still comprehensible and does not undermine the meaning of the original sentence. Similarly, ``Norwich'' and ``Southampton'' are both the name of cities in the England in the second sample, and ``frost'' and ``snow'' are semantically very similar words in the third example.

\begin{table*}[t!]
	\begin{center}
		\caption{Examples of language modeling outputs generated by RNN--VAE and RNN--SVAE} \label{table4.A few examples of language modeling outputs produced by VAE and SVAE}%
		\renewcommand{\arraystretch}{1.3}
		\small{
			\centering{ \setlength\tabcolsep{11pt}
				\begin{tabular}{c|l}
					\hline
					\textbf{Type} & \textbf{Sentence}\\ 
					\hline
					\textbf{Truth} & What is the name of the pension plan\\ 
					\hline
					\textbf{RNN--VAE} & \colorbox{black!20}{What is the name} and the name expires\\ 
					\hline
					\textbf{RNN--SVAE} & \colorbox{black!20}{What is the name of the pension plan} \\ 
					\hline
					\hline
					\textbf{Truth} & \makecell[l]{The grandmother of three who whishes to remain anonymous said the \\ experience was so traumatic she will never be able to eat popcorn again} \\
					\hline
					\textbf{RNN--VAE} & \makecell[l]{\colorbox{black!20}{The grandmother of} the kitchens of how serving ladies advised were the \\ added experience so you so fully flowers to likely \colorbox{black!20}{to eat}never desire}\\
					\hline 
					\textbf{RNN--SVAE} &  \makecell[l]{\colorbox{black!20}{The grandmother of three who whish to remain anonymous said the} \\ \colorbox{black!20}{experience was so traumatic she will never be able to eat} before pets \colorbox{black!20}{again}} \\ 		
					\hline
					\hline
					\textbf{Truth} & \makecell[l]{The authorities arrested two people but failed to investigate reports that they \\ were part of a large private militia}\\ 
					\hline
					\textbf{RNN--VAE} & \makecell[l]{\colorbox{black!20}{The authorities} raped some \colorbox{black!20}{people but failed to investigate reports that they}\\ \colorbox{black!20}{were part of a large private militia}}\\
					\hline
					\textbf{RNN--SVAE} &  \makecell[l]{\colorbox{black!20}{The authorities arrested two people but failed to investigate reports that they} \\\colorbox{black!20}{were part of a large private} Kuwaiti}\\  
					\hline
		\end{tabular}}}
	\end{center}
\end{table*}
\begin{table*}[t!]
	\begin{center}
		\caption{Examples of wrong words but have high similarity with the ground truth} \label{table5.Examples of wrong words but have high similarity with ground truth}%
		\renewcommand{\arraystretch}{1.3}
		\small{
			\centering{ \setlength\tabcolsep{5pt}
				\begin{tabular}{c|l}
					\hline
					\textbf{Type} & \textbf{Sentence}\\ 
					\hline
					\textbf{Truth} & \makecell[l]{These \underline{nodes} range from opening and closing tags to character data and processing \\ instructions}\\ 
					\hline
					\textbf{RNN--SVAE} &  \makecell[l]{These \underline{fibres} range from opening and closing tags to character data and processing \\ instructions}\\ 
					\hline
					\hline
					\textbf{Truth} & From there Jennings took the controls and flew to \underline{Norwich}\\ 
					\hline
					\textbf{RNN--SVAE} &  From there Jennings took the controls and flew to \underline{Southampton}\\ 
					\hline
					\hline
					\textbf{Truth} & \makecell[l]{Kevin Walker head of science at the \underline{BSBI} said the trend was down to the mild winter \\ and a lack of \underline{frost}}\\ 
					\hline
					\textbf{RNN--SVAE} &  \makecell[l]{Kevin Walker head of science at the \underline{UNK} said the trend was down to the mild winter \\ and a lack of \underline{snow}}\\ 
					\hline
		\end{tabular}}}
	\end{center}
\end{table*}
Figure \ref{Fig:BLEU} shows the average BLEU score of each model per the sentence length of each dataset. When the sentence length is relatively short, sentence generation performance was similar. Even the RNN--AE worked well with very short sentences (i.e., less than five words). Additionally, there was no significant difference between the RNN--VAE and RNN--SVAE. When the sentence length was moderate, RNN--AE tended to fail to generate the original sentence; its BLEU scores were much lower than those of the other two methods. When comparing RNN--VAE and RNN--SVAE, the RNN--SVAE worked better than RNN--VAE in most cases. When the sentence length was long, all three models had trouble generating the original sentence. This is still an open research topic in the field of machine translation.

\subsection{Missing Word Imputation}
Missing word imputation is the process of completing a sentence by filling it in with appropriate words \cite{manisolving}. We performed this task to evaluate how well the proposed RNN--SVAE reflects the global latent feature of the input sentences. In this task, an incomplete sentence with some words erased was provided as an input to the encoder of seq2seq models. The models were trained to guess the erased words or the sequence of words correctly through the decoder. We tested the missing word imputation performance under three different scenarios. The detailed description of each scenario is summarized below.

\begin{itemize}
	\item \textbf{Scenario 1}: Imputation for the last word of the sentence.
	\item \textbf{Scenario 2}: Imputation for one randomly selected word among the last 20\% of the sentence.
	\item \textbf{Scenario 3}: Imputation for the sequence of words corresponding to the last 20\% of the sentence.
\end{itemize}
Scenario 1 was the easiest level and Scenario 3 was the most difficult level. Scenario 1 and 2 can be regarded as a multi-class classification task, whereas Scenario 3 can be regarded as a sequence generation task.
\begin{table*}[t!]
	\begin{center}
		\caption{Performance of missing word imputation task.} \label{table6.Result of missing word imputation task}%
		\renewcommand{\arraystretch}{1.3}
		\small{
			\centering{ \setlength\tabcolsep{20pt}
				\begin{tabular}{c|c|c|c}
					\hline
					\backslashbox{Model}{Scenario} &  \makecell{Scenario 1\\(Accuracy)}&  \makecell{Scenario 2\\(Accuracy)}& \makecell{Scenario 3\\(BLEU)}\\ 
					\hline
					RNN--AE &  15.71 &  5.76 &  34.08 \\ 
					\hline
					RNN--VAE & \textbf{$$16.94$$} & 6.23 & 34.17\\ 
					\hline
					RNN--SVAE & 15.05 & \textbf{$$6.37$$} & \textbf{$$34.37$$}\\ 
					\hline
		\end{tabular}}}
	\end{center}
\end{table*}

\begin{table*}[t!]
	\begin{center}
		\caption{Examples of missing word imputation.} \label{table7.Several examples of missing word imputation}%
		\renewcommand{\arraystretch}{1.3}
		\small{
			\centering{ \setlength\tabcolsep{4pt}
				\begin{tabular}{c|c|c|c|c}
					\hline
					& \multicolumn{4}{c}{\makecell[l]{Q: Inventories increased across divisions, but were compensated by \\
							advance payments received and a better operational $\rule{1cm}{0.15mm}$.}} \\\cline{2-5}
					&\cellcolor{black!30}\textbf{Truth} & \cellcolor{black!30}\textbf{RNN--AE} & \cellcolor{black!30}\textbf{RNN--VAE} & \cellcolor{black!30}\textbf{RNN--SVAE} \\\cline{2-5}
					& “performance” & “performance” & “performance” & “service” \\\cline{2-5}
					
					&\multicolumn{4}{c}{\makecell[l]{Q: Click here for instructions on how to enable javascript in your $\rule{1cm}{0.15mm}$.}} \\\cline{2-5}
					&\cellcolor{black!30}\textbf{Truth} & \cellcolor{black!30}\textbf{RNN--AE} & \cellcolor{black!30}\textbf{RNN--VAE} & \cellcolor{black!30}\textbf{RNN--SVAE} \\\cline{2-5}
					\multirow{-7}{*}{\textbf{Level 1}} 
					& “browser” & “browser” & “browser” & “system” \\
					\hline
					& \multicolumn{4}{c}{\makecell[l]{Q: David Rhodes and Robert Hendricks (Montreal process technical advisory \\ group tac) described tac’s work on a framework of criteria and indicators that \\ provide a common of  $\rule{1cm}{0.15mm}$ management of temperate and boreal forests.}} \\\cline{2-5}
					&\cellcolor{black!30}\textbf{Truth} & \cellcolor{black!30}\textbf{RNN--AE} & \cellcolor{black!30}\textbf{RNN--VAE} & \cellcolor{black!30}\textbf{RNN--SVAE} \\\cline{2-5}
					& “sustainable” & “the” & “the” & “sustainable” \\\cline{2-5}
					
					&\multicolumn{4}{c}{\makecell[l]{Q: Such distinctive homes can attract interest from far beyond your $\rule{0.7cm}{0.15mm}$ market.}} \\\cline{2-5}
					&\cellcolor{black!30}\textbf{Truth} & \cellcolor{black!30}\textbf{RNN--AE} & \cellcolor{black!30}\textbf{RNN--VAE} & \cellcolor{black!30}\textbf{RNN--SVAE} \\\cline{2-5}
					& “local” & “local” & “local” & “own” \\\cline{2-5}
					
					&\multicolumn{4}{c}{\makecell[l]{Q: The list is sorted by country so you shouldn’t have a problem to find a $\rule{0.7cm}{0.15mm}$\\ near you.}} \\\cline{2-5}
					&\cellcolor{black!30}\textbf{Truth} & \cellcolor{black!30}\textbf{RNN--AE} & \cellcolor{black!30}\textbf{RNN--VAE} & \cellcolor{black!30}\textbf{RNN--SVAE} \\\cline{2-5}
					\multirow{-11}{*}{\textbf{Level 2}}
					& “vendor” & “destination” & “few” & “hotel” \\
					\hline
					
					& \multicolumn{4}{c}{\makecell[l]{Q: If you have text in any page of your site that contain any of the keywords \\ below, you can add your contextual listing there. It’s free and your listing will \\ appear online in $\rule{0.7cm}{0.15mm}$. }} \\\cline{2-5}
					&\cellcolor{black!30}\textbf{Truth} & \cellcolor{black!30}\textbf{RNN--AE} & \cellcolor{black!30}\textbf{RNN--VAE} & \cellcolor{black!30}\textbf{RNN--SVAE} \\\cline{2-5}
					& \makecell{“real time conta-\\ining hyperlink \\ to your page”} & \makecell{“real time http \\ www ‘UNK’ com \\ au account”} & \makecell{“real time conta-\\ining your account \\ to your account}” & \makecell{“real time conta-\\ining hyperlink \\ to your page”} \\\cline{2-5}
					
					& \multicolumn{4}{c}{\makecell[l]{Q: Energy star is a registered trademark of the US environmental $\rule{0.7cm}{0.15mm}$. }} \\\cline{2-5}
					&\cellcolor{black!30}\textbf{Truth} & \cellcolor{black!30}\textbf{RNN--AE} & \cellcolor{black!30}\textbf{RNN--VAE} & \cellcolor{black!30}\textbf{RNN--SVAE} \\\cline{2-5}
					& “protection agency” & “protection agency” & “protection agency” & “insurance program” \\\cline{2-5}
					
					&\multicolumn{4}{c}{\makecell[l]{Q: Encrypt within the veritas net backup policy eliminating a seperate process \\ or an extra dedicated $\rule{0.7cm}{0.15mm}$.}} \\\cline{2-5}
					&\cellcolor{black!30}\textbf{Truth} & \cellcolor{black!30}\textbf{RNN--AE} & \cellcolor{black!30}\textbf{RNN--VAE} & \cellcolor{black!30}\textbf{RNN--SVAE} \\\cline{2-5}
					
					\multirow{-12}{*}{\textbf{Level 3}}
					& \makecell{“device to manage”} & \makecell{“to the application”} & \makecell{“to the enviroment}” & \makecell{“device to manage”} \\
					\hline
		\end{tabular}}}
	\end{center}
\end{table*}
\subsubsection{Data Set}
For model training, the training dataset used in the language modelling task (i.e., the randomly sampled 2,500,000 sentences from the News Crawl'13 dataset) was modified. Likewise, we modified the News Crawl'13 test dataset and used it to evaluate performance. For Scenarios 1 and 3, we erased the last word and the last 20\% of word sequences from each sentence, respectively. For Scenario 2, we replaced a randomly selected word among the last 20\% of the sentence with $0$ vector. 
\begin{figure*}
	\centering
	\includegraphics[width=0.6\textwidth]{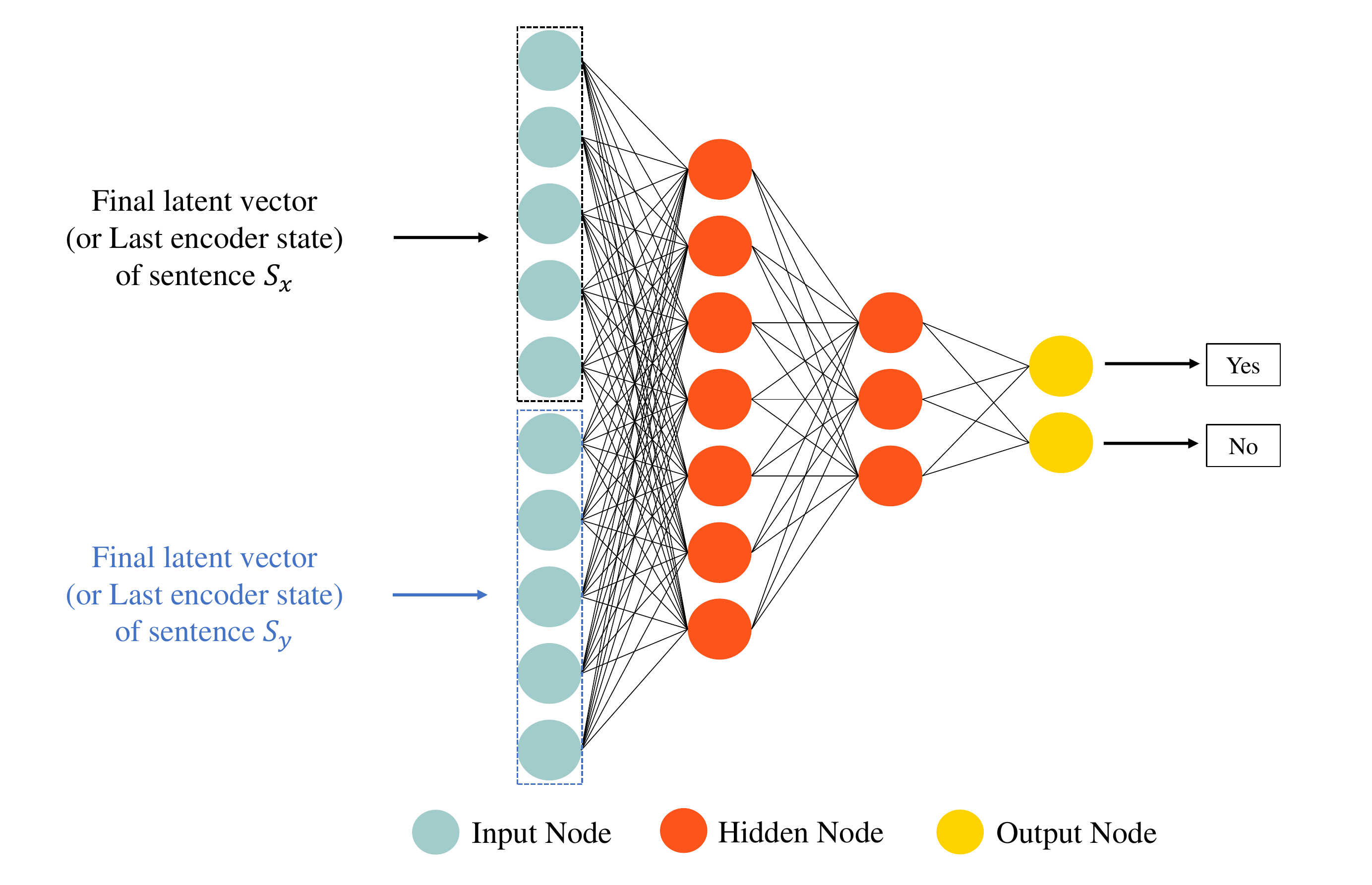}
	\caption{Structure of paraphrase identification model}
	\label{fig:pi_structure}
\end{figure*}
\subsubsection{Model Training and Inference}
Three imputation models were trained under the same condition. We used the Xavier initialization for parameter initialization and the Adam optimizer. The models were trained for 15 epochs with a learning rate 0.001 for the first five epochs and 0.0001 for remaining 10 epochs. Gradient computation and weight updates were done with a mini-batch size of 512. Like the language modeling task, the outputs of RNN--VAE and RNN--SVAE were decoded from the mean vector of five sampled values to reduce the bias of the global latent vector.

\subsubsection{Results}
As a quantitative evaluation metric, the simple accuracy (i.e., the proportion of the correctly predicted words to the number of total missing words) was used for Scenarios 1 and 2 (i.e., predicting a single word), whereas the BLEU score was used for Scenario 3 (i.e., predicting a sequence of words). 

Table \ref{table6.Result of missing word imputation task} shows the performance of each model for missing word imputation. For Scenario 1, RNN--VAE yielded the highest accuracy, whereas RNN--SVAE resulted in the lowest accuracy. Because imputation for the last word requires more information about the end of the sentence than the global information of the whole sentence, RNN--VAE and RNN--AE, which preserve more information of the end of sentences, showed good performances. Although RNN--SVAE resulted in the worst performance, we found that its imputation results were semantically quite similar to the target word in many examples, as shown in Table \ref{table7.Several examples of missing word imputation}. 

For more difficult tasks, such as Scenario 2 and 3, on the other hand, the RNN--SVAE outperformed the other methods. As shown in Table \ref{table7.Several examples of missing word imputation}, not only did RNN--SVAE achieve higher accuracy, or BLEU score, it also predicted semantically similar words to the correct answers.

\subsection{Paraphrase Identification}

Paraphrase identification is a task that determines whether two different sentences have the same meaning \cite{rus2008paraphrase,hu2014convolutional}. In this study, we constructed a binary classification model to determine whether two sentences are paraphrased when global latent vectors, or the last hidden state for RNN—AE, of each sentence are used as input, as shown in Figure \ref{fig:pi_structure}. Like the previous tasks, the mean vector of five sampled vectors is used as input to the paraphrase identification model for RNN--VAE and RNN--SVAE. The model is constructed with a feed-forward multi-layer perceptron consisting of two hidden layers. The number of hidden units of the first and the second layer are set to 100 and 50, respectively.

\begin{table*}[t!]
	\begin{center}
		\linespread{1.3}
		\caption{Result of paraphrase identification} \label{table8.Result of paraphrase identification}%
		\small{
			\centering{ \setlength\tabcolsep{13pt}
				\begin{tabular}{c|c|c|c}
					\hline
					\backslashbox{Model}{Metric} 
					&\makecell{Error rate\\(1 - Accuracy)}
					&\makecell{False alarm rate \\(1 - Precision)}
					&\makecell{Miss rate\\(1 - Recall)}\\ 
					\hline
					RNN--AE &  5.10 $\pm$ 0.56 &  5.09 $\pm$ 0.80 &  \textbf{5.59} $\pm$ \textbf{0.92} \\
					\hline
					RNN--VAE & 6.05 $\pm$ 0.43 & 5.64 $\pm$ 0.78 & 6.02 $\pm$ 0.88\\ 
					\hline
					RNN--SVAE & \textbf{4.65} $\pm$ \textbf{0.33} & \textbf{3.68} $\pm$ \textbf{0.65} & 5.86 $\pm$ 0.71\\ 
					\hline
		\end{tabular}}}
	\end{center}
\end{table*}

\begin{table*}[t!]
	\begin{center}
		\linespread{1.3}
		\caption{Result of paraphrase sentence similarity} \label{table9.Result of paraphrase sentence similarity}%
		\small{
			\centering{ \setlength\tabcolsep{30pt}
				\begin{tabular}{c|c|c}
					\hline
					Model & RNN-VAE & RNN--SVAE \\
					\hline
					Cosine similarity & 0.598 & \textbf{0.631} \\ 
					\hline
		\end{tabular}}}
	\end{center}
\end{table*}

\subsubsection{Data Set}
We used the MS Paraphrase Corpus dataset \cite{dolan2004unsupervised,quirk2004monolingual} to perform the paraphrase identification task. This dataset consists of 5,801 pairs of sentences with 4,076 pairs for training and 1,725 pairs for test. The training dataset consists of 2,753 ``equivalent'' sentence pairs and 1,323 ``not equivalent'' sentence pairs, as judged by human raters. The test set consists of 1,147 and 578 ``equivalent'' and ``not equivalent'' sentence pairs, respectively.

\citet{dolan2004unsupervised} noted that, although the collected paraphrase sentences were judged ``not equivalent'' by the human raters, it was not desirable to use ``not equivalent'' sentence pairs as negative class data, because they have significant overlaps between them, in terms of information content and wording. Therefore, we used "equivalent" sentences of the MS Paraphrase Corpus as the positive class dataset and modified one side of the sentence pair to use as the non-paraphrase dataset. The non-paraphrase dataset is generated by replacing 20\% of randomly selected words in a paired sentence with other words in the pre-trained word vector dictionary used in language modeling and missing word imputation tasks. For the training data, we used 2,753 pairs of sentences as the positive class and generated 2,753 pairs of negative class sentences by using the method described above. Similarly, 1,147 pairs of sentences for the test were used as the positive class, and 1,147 pairs of negative class sentences were generated for the test data. Thus, a total of 5,506 training pairs and 2,294 test pairs were constructed. The ratio of paraphrase pairs to non-paraphrase pairs was the same.

\subsubsection{Training Details}
The paraphrase identification models for RNN--AE, RNN--VAE, and RNN--SVAE were trained under the same conditions. The parameters of all models were initialized by using Xavier initialization. Gradient computation and weight updates were done with a mini-batch size of 512. The models were trained for 100 epochs using the Adam optimizer with a learning rate of 0.001. To prevent overfitting, dropout \cite{srivastava2014dropout} is used for each layer. The dropout rate is set to 0.3. We repeated training 30 times for each model to obtain the statistical significance of the results.

\subsubsection{Results}
We used three evaluation metrics: (1) the overall error rate, (2) false alarm rate (i.e., the proportion of incorrectly classified paraphrases as ``equivalent'' among the paraphrases classified as ``equivalent'' by the model) and (3) miss rate (i.e., the proportion of incorrectly classified paraphrases that were actually ``equivalent'' among the actual ``equivalent'' paraphrases). The results and standard deviations of paraphrase identification task of each model are summarized in Table \ref{table8.Result of paraphrase identification}. The RNN--SVAE resulted in better performance than RNN--AE and RNN—VAE in terms of error rate and false alarm rate. These performance improvements are also supported by the statistical hypothesis testing at a significant level of 0.01. Although RNN--AE showed the best performance in terms of miss rate, there is no statistically significant difference between the performance of RNN--AE and that of RNN--VAE or RNN--SVAE at a significant level of 0.01. Compared to RNN--VAE, RNN--SVAE reduced the error rate by 23.1\% $\sim$ 34.8\%, which strongly supports the notion that the RNN--SVAE can better capture the global latent context over RNN--VAE.

In addition to paraphrase identification, which was evaluated by the binary decision, we also compared the similarity between latent vectors of two ``equivalent'' sentences judged by human raters. This evaluation was conducted only with RNN--VAE and RNN--SVAE to exploit the effect of adding document information vector to variational-based RNN models. Table \ref{table9.Result of paraphrase sentence similarity} shows that not only did the RNN--SVAE model achieve higher identification accuracy, it also generated more similar latent vectors for two similar sentences than RNN--VAE.

\section{Conclusion}
For RNN--based autoencoder models (e.g., RNN--AE and RNN—VAE) the final hidden state of the encoder does not contain sufficient information about the entire sentence. In this paper, we proposed RNN--SVAE to overcome this limitation. To consider the information of words in the sentence, we constructed a document information vector by a linear combination of word vectors of input sentence. The weights of individual words are computed using the attention information between the final state of the encoder and every prior hidden state. We then combined this document information vector with the final hidden state of the bi-directional RNN encoder to construct the global latent vector as the output of the encoder part. Then, the mean and standard deviation of the continuous semantic space were learned to take advantage of variational method.

The proposed RNN--SVAE was verified through three NLP tasks: language modeling, missing word imputation, and paraphrase identification. Despite the simple structure of RNN--SVAE combining the document information vector with the RNN--VAE model, experimental results showed that RNN--SVAE achieved higher performance than RNN--AE and RNN--AE for all tasks requiring global latent meaning of the input sentence. The only exception is missing word imputation for a very short sentence, which does not significantly depend on the global semantic information. 

Although the experimental results are very favorable for RNN--SVAE, there are some limitations of the current study. This provides some future research directions. First, the prior distribution is assumed to be a specific distribution, such as standard Gaussian. To improve the performance of RNN--SVAE, it will be worth attempting to find an appropriate prior distribution of data. Additionally, there is the risk of learning a model that is far from the actual data distribution. Thus, as in adversarial autoencoder \cite{makhzani2015adversarial} for image data, further research is needed to map prior distribution to data distribution in language modeling. Second, we should use the Bi-RNN structure to find the weight of a word that is not biased on one side of the sentence. To apply RNN--SVAE to one-directional RNN structures, it is necessary to study a method of re-adjusting weight properly, so that the weight of words is not biased to one side.

\bibliographystyle{./sty/icml2017} 
\bibliography{References}
\end{CJK}
\end{document}